\documentclass[letterpaper, 10 pt]{IEEEtran}  %

\IEEEoverridecommandlockouts
\usepackage{cite}
\usepackage{amsmath,amssymb,amsfonts}
\usepackage{algorithmic}
\usepackage{graphicx}
\usepackage{textcomp}
\usepackage{xcolor}
\usepackage{balance}
\usepackage{comment}
\usepackage{caption}
\usepackage{array}
\usepackage{subcaption}
\usepackage{arydshln}
\usepackage{multirow, tabularx}
\usepackage{tensor}
\usepackage{hyperref}
\usepackage{float}
\usepackage{rotating}
\usepackage{nicematrix}
\usepackage{mathtools}
\usepackage{amsmath}
\usepackage{pifont}
\usepackage{cuted}
\usepackage{threeparttable}
\usepackage{graphicx}
\usepackage{amsmath}
\usepackage{amssymb}
\usepackage{booktabs}
\usepackage{enumitem}

\usepackage[linesnumbered,ruled,vlined]{algorithm2e}
\SetKwInput{KwInput}{Input}                
\SetKwInput{KwOutput}{Output}              

\usepackage{subcaption}
\usepackage{xcolor}
\usepackage{colortbl}
\definecolor{cGreen}{RGB}{0,255,0}
\definecolor{LightCyan}{rgb}{0.88,1,0.88}

\newcommand{\eg}{\emph{e.g.},}
\newcommand{\ie}{\emph{i.e.},}
\newcommand{\etal}{\emph{et~al.}}
\definecolor{LightCyan}{rgb}{0.88,1,0.88}
\definecolor{emb_color}{RGB}{252,224,225}
\definecolor{multi_head_attention_color}{RGB}{252,226,187}
\definecolor{add_norm_color}{RGB}{242,243,193}
\definecolor{ff_color}{RGB}{194,232,247}
\definecolor{softmax_color}{RGB}{203,231,207}
\definecolor{linear_color}{RGB}{220,223,240}
\definecolor{gray_bbox_color}{RGB}{243,243,244}

\def \pgat {P-GAT}
\def \minkloc {MinkLoc3D }

\def \superglue {SuperGlue }
\def\secref#1{Sec.~\ref{#1}}
\def\figref#1{Fig.~\ref{#1}}
\def\tabref#1{Table~\ref{#1}}
\def\eqref#1{Eq.~(\ref{#1})}
\def\algref#1{Alg.~\ref{#1}}

\definecolor{LightCyan}{rgb}{0.88,1,0.88}

\def\BibTeX{{\rm B\kern-.05em{\sc i\kern-.025em b}\kern-.08em
    T\kern-.1667em\lower.7ex\hbox{E}\kern-.125emX}}
\begin{document}

\title{\LARGE \bf Pose-Graph Attentional Graph Neural Network\\ for Lidar Place Recognition
}
\author{Milad Ramezani$^{1{\ast}{\dagger}}$, Liang Wang$^{1,2{\dagger}}$, Joshua Knights$^{1,3}$, Zhibin Li$^{1}$, Pauline Pounds$^{2}$, Peyman Moghadam$^{1,3}$
\thanks{
$^1$ Robotics and Autonomous Systems, DATA61, CSIRO, 
Australia. 
E-mails: {\tt\footnotesize \emph{firstname.lastname}@csiro.au}}
\thanks{
$^{2}$ The University of Queensland, Brisbane, Australia.
E-mails: {\tt\footnotesize \emph{firstname.lastname}@uq.edu.au}
}
\thanks{
$^{3}$
Queensland University of Technology (QUT), Brisbane, Australia.
E-mails: {\tt\footnotesize \emph{firstname.lastname}@qut.edu.au}}} 

\maketitle
\def\thefootnote{$\ast$}\footnotetext{Corresponding author}
\def\thefootnote{$\dagger$}\footnotetext{These authors contributed equally to this work.}

\begin{abstract}

This paper proposes a pose-graph attentional graph neural network, called \pgat, which compares (key)nodes
between sequential and non-sequential sub-graphs for place recognition
tasks as opposed
to a common frame-to-frame retrieval problem formulation
currently implemented in SOTA place recognition methods. \pgat~uses the maximum spatial and temporal information between neighbour cloud descriptors \----generated by an existing encoder\---- utilising the concept of pose-graph SLAM.
Leveraging intra- and inter-attention and graph neural network,~\pgat~relates point clouds captured in nearby locations in Euclidean space and their embeddings in feature space. 
Experimental results on the large-scale publically available datasets demonstrate the effectiveness of our approach in scenes lacking distinct features and 
when training and testing environments have different distributions (domain adaptation). Further, an exhaustive comparison with the state-of-the-art shows improvements in performance gains.   
Code is available at \url{https://github.com/csiro-robotics/P-GAT}.  

\end{abstract}

\section{Introduction}
\label{sec:intro}

\IEEEPARstart{A}{ccurate} and drift-free (re)-localisation is critical for many robotic and computer vision applications, such as autonomous navigation~\cite{huang2023fael} and augmented reality~\cite{sarlin2022lamar}. Achieving reliable (re)-localisation is challenging, particularly in GPS-denied environments, such as indoor, subterranean or dense vegetated environments~\cite{knights2023wildplaces, ramezani2023deep}, due to occlusion, complex geometry, and dynamic objects in scenes. 

One promising direction for addressing the challenges associated with reliable (re)-localisation is to utilise a Place Recognition (PR) method to predict the coarse location of an agent within a database of previously visited places.  Place Recognition is commonly framed as a retrieval task in computer vision and robotics, either vision-based~\cite{wang2022transvpr, ali2023mixvpr, zhang2023etr, li2023hot} or lidar-based~\cite{uy2018pointnetvlad, jacek20minkloc, zhang2019pcan, xia2021soe, hui2021pyramid}. Given a query, the method involves retrieving the most similar key in the database by first encoding the input frame (image/point cloud) as a global descriptor and matching it against the global descriptors of previously visited places. 
\begin{figure}
    \centering
    \includegraphics[width=0.9\linewidth]{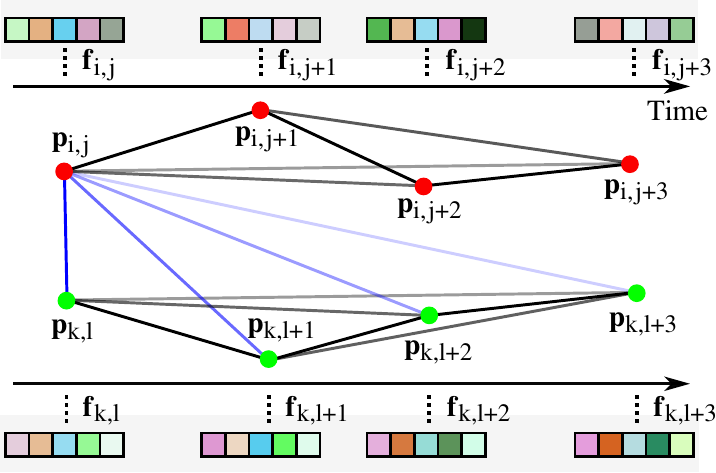}
    \caption{\small{\pgat~aims to optimise spatiotemporal information by relating point clouds within and across subgraphs leveraging an attentional graph neural network. If point clouds are captured in nearby locations (similar point clouds), our intra- (edges in black shades) and inter- (edges in blue shades) attention mechanism reweights their embeddings to bring them closer in feature space. Intra-attention enhances distant point cloud communication within subgraphs, while inter-attention facilitates potential place recognition in revisit areas.}}
\label{fig:teaser}
\vspace{-0.5cm}
\end{figure}

Despite remarkable improvements, visual PR is less robust against appearance, season, illumination and viewpoint variations in large-scale (\ie{} city-scale) areas. In this paper, we consider the problem of lidar place recognition for large-scale environments. Despite all the progress in the field of lidar place recognition, most existing methods only encode a single lidar frame into a global descriptor, and hence, topological scene-level understanding is often neglected. There are few prior works \cite{ma2022seqot, vidanapathirana2021locus} that directly aggregate a sequence of lidar descriptors to generate one single global descriptor for each lidar sequence. However, these methods do not take advantage of the topological relationship between a sequence of point clouds in the context of a graph containing sets of nodes and edges. %

To exploit the spatiotemporal information between neighbouring point clouds, we propose an attentional graph neural network called \pgat, that uses topological information obtained by a pose-graph lidar SLAM system, between a set of point clouds to maximise the receptive field for training. 
Nodes and edges generated in a pose graph (as the robot explores the environment and pose-graph SLAM optimises robot poses) are further used to generate fully connected graphs (subgraphs) that contain positional information for a given robot travel distance.
These subgraphs are further fed into our \pgat~model for place recognition by comparing a pair of subgraphs rather than just a pair of point clouds (\figref{fig:teaser}).  
The communication between the nodes of the subgraph pairs is performed leveraging an attentional graph neural network.
To address the dynamic properties of subgraphs (varying number of nodes) and the presence of nodes in multiple subgraphs (in contrast to comparing between pairs of point clouds/images), we develop a customised layer normalisation module using a boolean mask for padding nodes and develop an averaging scheme for efficiently computing similarity scores during inference.
We demonstrate that \pgat~can be integrated into any global point cloud descriptors (\ie~cloud encoder agnostic), which greatly improves their robustness and generalisability. We extensively analyse and compare the performance of our network with the state-of-the-art over multiple large-scale public datasets. 
To characterise the properties of \pgat{} in detail, we demonstrate the role of each component using numerous ablation studies.
The proposed \pgat~can achieve the state-of-the art on various benchmark datasets for lidar place recognition tasks.

\section{Related Work}
\label{sec:related}
This section reviews hand-crafted and learning-based lidar PR methods before introducing the attentional graph neural network and its applications. 

\subsection{Handcrafted Lidar Descriptors}
The purpose of lidar PR methods is to create distinctive descriptors, based on a Local Reference Frame (LRF), along the robot's path to recognise revisited places regardless of its pose. Two types of descriptors are used: signatures and histograms. Histogram-based methods such as PFH~\cite{rusu2008aligning} and FPFH~\cite{rusu2009fast}, describe the 3D surface neighbourhood of a point by encoding a few geometric features obtained individually at each point according to local coordinates. DELIGHT~\cite{cop2018delight}, instead of geometric properties, computes intensity histograms. Signature-based algorithms like Scan Context~\cite{kim2018scan} or \textit{Segmatch}~\cite{dube2017segmatch}  use descriptor-based features to improve place recognition. Additionally, algorithms such as SHOT~\cite{salti2014shot} combine both histograms and signatures for robust place recognition.
However, handcrafted lidar descriptors require careful tuning of feature extraction and matching parameters depending on the operating environment.

\subsection{Learning-based Lidar Descriptors} 
Recent advances in learning-based approaches have shown promising results in addressing the challenges mentioned earlier. CNN-based methods such as \textit{Segmap}~\cite{dube2018segmap} or Efficient Segment Matching (ESM)~\cite{tinchev2019learning} encode local patches of a point cloud into local embeddings. Local descriptors can later be used for localisation~\cite{dube2020segmap, ramezani2020online}. These methods, however, are not defined in an end-to-end fashion to form a scene-level global descriptor for point clouds. 

Using a convolutional bottom-up and top-down backbone,  MinkLoc3D~\cite{jacek20minkloc} and its variations~\cite{zywanowski2021minkloc3d, komorowski2021egonn} extract local features and aggregate them into a global descriptor by a Generalised-Mean pooling (GeM)~\cite{lin2017feature}.~\textit{LoGG3D-Net}~\cite{vidanapathirana2022logg3d} employs a sparse convolutional U-Net to encode point clouds into local features. During training, it ensures the maximum similarity of corresponding local features on a pair of neighbour point clouds by defining a local consistency loss. Unlike MinkLoc3D, LoGG3D-Net uses second-order pooling to aggregate local features to create global descriptors.

In contrast to convolutional models, methods have been proposed which are based on PointNet~\cite{qi2017pointnet}, an encoder which works directly on an unordered point cloud due to permutation invariance to points utilising a symmetry function. PointNetVLAD~\cite{uy2018pointnetvlad} is a seminal lidar PR work with a PoinNet-based backbone design. It uses NetVLAD~\cite{arandjelovic2016netvlad} to aggregate local descriptors for the generation of global descriptors. To capture local contextual information, PCAN~\cite{zhang2019pcan} adds an attention map mechanism for predicting the significance of point features. Re-weighted local features are further aggregated into a discriminative global descriptor using NetVLAD. Similarly, LPD-Net~\cite{liu2019lpd} aims to cover the limitation of PointNet in the extraction of local contextual information by aggregating neighbour features using a graph neural network. Global descriptors are finally generated using NetVLAD. In another effort, SOE-Net~\cite{xia2021soe} adds an orientation-encoding unit in local descriptor extraction and a self-attention unit before the aggregation of local features through NetVLAD to improve the point-wise feature representation of PointNet. Recently, PPT-Net~\cite{hui2021pyramid}, inspired by SOE-Net~\cite{xia2021soe} and pyramid structure of PointNet++~\cite{qi2017pointnet++}, proposed a pyramid point transformer design to learn the regional contextual information of a point cloud at multiple levels leveraging a grouped self-attention mechanism for the extraction of discriminative local embeddings. Local embeddings are also aggregated using NetVLAD. 

In all the lidar PR methods mentioned above, point clouds are compared pairwise and thus, the topological and sequential relationship between a set of point clouds are not explored.  The work that attempts to leverage temporal information between point clouds is Locus~\cite{vidanapathirana2021locus} which aims to relate the local features of the current point cloud to their correspondences in the previous point clouds using second-order temporal feature pooling; however, within a short time window consisting of only three frames. This limits the generalisation of Locus~\cite{vidanapathirana2021locus} under test-time distribution shifts. SeqOT~\cite{ma2022seqot} was also proposed to benefit from temporal information, although using a sequence of range images. Moreover, it generates only a single descriptor for each sequence.

\subsection{Attentional Graph Neural Network} 
Attentional Graph Neural Network (AGNN) 
is designed to process data represented as a graph, using attention mechanisms~\cite{vaswani2017attention} to selectively focus on nodes and edges. The attention mechanism helps reduce the graph's complexity and improves the effectiveness of local feature matching across nodes.  %
AGNNs have been successfully applied to various tasks, such as image matching~\cite{sarlin20superglue}, object detection~\cite{min2021attentional}, and program re-identification~\cite{wang2019attentional}. To the best of our knowledge, AGNN has never been used in lidar PR tasks. We believe AGNN is well-suited for lidar PR (due to its efficiency in the aggregation of contextual information) when exploring spatiotemporal information existing between nearby point clouds. 

\section{The Proposed Method}
\label{sec:method}

\begin{figure*}[t]
    \centering
    \includegraphics[width=0.9\linewidth]{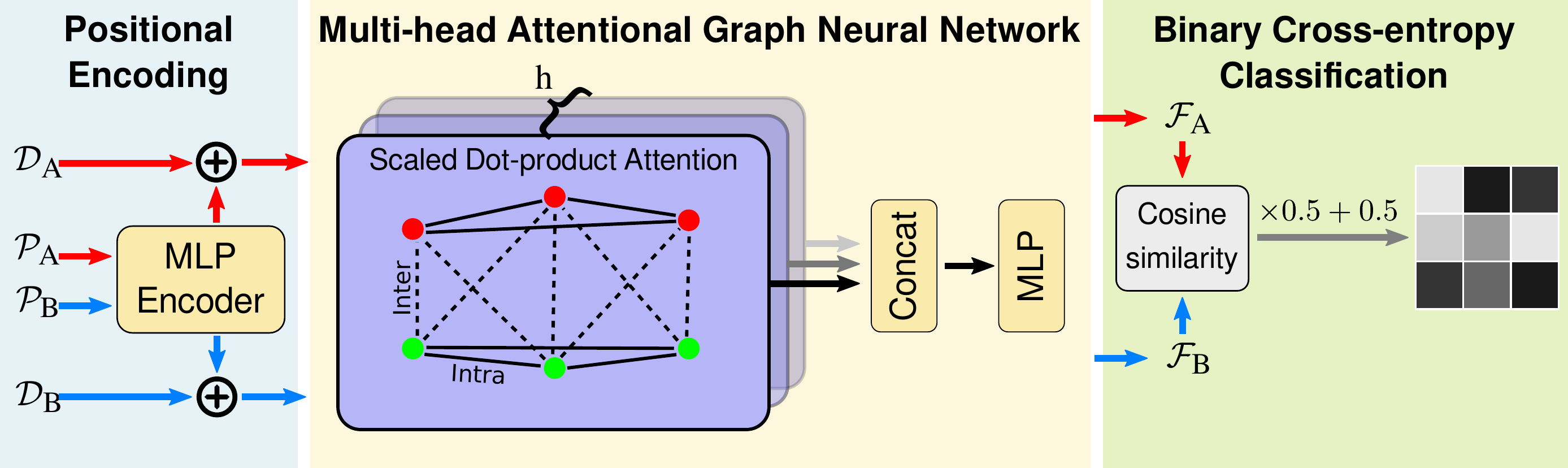}
    \caption{\small{\pgat~employs a multi-head attentional graph neural network for contextual and viewpoint information aggregation of sequential point clouds, refining each point cloud's descriptor. The network uses an MLP encoder that supplements descriptors with normalised positions. Intra- and inter-attention mechanisms infer relationships for all nodes simultaneously, updating the descriptors accordingly.
    \pgat~predicts whether the pair of two point clouds are captured from the same place by comparing their final descriptors with shifted cosine similarity. 
    }}
\label{fig:architecture}
\vspace{-0.2cm}
\end{figure*}
Our goal is to enhance PR performance by increasing the spatiotemporal information on scenarios where a graph with nodes and edges represents robot poses and spatial constraints in between. Robot poses serve as positional information. 

\subsection{Overview}
\pgat~is defined based on a graph $\mathcal{G} = (\mathcal{V}, \mathcal{E})$, where $\mathcal{V}$ and $\mathcal{E}$ denote nodes and edges, respectively. The graph is inherently created by a SLAM system (pose-graph). For example, a lidar SLAM system registers consecutive lidar scans to create nodes and edges of the pose graph for further use in a back-end optimisation problem.  
Utilising pose-graph SLAM, we assume that accurate localisation is achievable within a local window of robot traversal,~\eg~200 m, for a vehicle travelling in a large-scale environment. However, no global localisation is provided to \pgat. 

Our network comprises three major blocks: positional encoding, graph-based attention mechanism and binary classification. Provided local robot poses by SLAM, we encode relative positional information into embeddings computed by a 3D backbone to maintain the topological relationship between a sequence of keynodes (selected robot poses associated with a point cloud and its corresponding descriptor). \pgat~utilises both graph structure and attention mechanism, described in~\secref{sec:agnn}, to increase the distinctiveness of positional-aware embeddings. Inspired by \superglue\cite{sarlin20superglue}, which uses an AGNN for image matching, we aim to relate the descriptors of point clouds captured in nearby locations by aggregating contextual and viewpoint information into point clouds' descriptors using an AGNN. Figure~\ref{fig:architecture} depicts the overall architecture of P-GAT and its components. 

\subsection{Problem Formulation}
To formulate the problem, we consider two sets of point clouds as two subgraphs $\mathcal{S}_A = \{\mathcal{V}^A_i\}_{i=1}^N$, and $\mathcal{S}_B = \{\mathcal{V}^B_j\}_{j=1}^M$, where $M$ and $N$ denote the number of keynodes in $\mathcal{S}_A$ and $\mathcal{S}_B$, respectively. Our PR problem is now defined between pairs of subgraphs instead of two individual point clouds, as is common in the literature. We also formulate the PR problem as a binary classification problem using a similarity measure computed from descriptors between every pair of $(\mathcal{V}^A_i, \mathcal{V}^B_j)$ in subgraphs $\mathcal{S}_A$ and $\mathcal{S}_B$. 

All the pairs of keynodes from subgraphs $\mathcal{S}_A$ and $\mathcal{S}_B$ that are captured in the same place must be classified. 
 As noted earlier, each keynode is associated with the robot position $\mathbf{t}$ and a descriptor $\mathbf{d}$,~\ie~$\mathcal{V} = \{\mathbf{t}, \mathbf{d}\}$. The position $\mathbf{t}\in \mathbb{R}^{3}$ is relative to the first node of the graph and normalised,~\ie~$\mathbf{p} = \frac{1}{\sigma}(\mathbf{t}-\mathbf{c})$, where \textbf{c} is the centroid of the nodes' positions in a subgraph and $\sigma$ is a scalar showing the scatteredness of the keynodes. This normalisation is essential to maintain the geometry consistency between keynodes and network generalisation when training and testing data have different distributions. The descriptor $\mathbf{d} \in \mathbb{R}^E$ represents the input point cloud embedding as a fixed-size vector global descriptor at the time of keynode $\mathcal{V}$, and $E$ denotes the descriptor dimension.

\subsection{Attentional Graph Neural Network}
\label{sec:agnn}
To exploit the spatiotemporal information existing between keynodes in the pose graph created by SLAM, we build a fully-connected graph between the subgraph pairs $(\mathcal{S}_A, \mathcal{S}_B)$,~\ie~$\mathcal{G}_{AB}=(\mathcal{V}_{AB},\mathcal{E}_{AB})$ with $N + M$ keynodes and $\frac{1}{2}((N + M)^2-(N + M))$ undirected edges. Now, we can aggregate the relative positional and contextual constraints between the keynodes in $\mathcal{G}_{AB}$ utilising an AGNN to enhance the descriptors' representation. Since $\mathcal{G}_{AB}$ has two types of connections (multiplexity~\cite{mucha2010community}),~\ie~intra- and inter- subgraph edges, we effectively train the network to push the embeddings of the point clouds captured in nearby locations together. This allows embeddings to be invariant to dynamic and viewpoint changes resulting in more effective place recognition.

\noindent
\textbf{Positional Encoding:} To embed the keynode relative position into the descriptor with a higher dimension $E$ in each subgraph, we use Multi-Layer Perceptron (MLP) encoder to increase the dimension of the position $\mathbf{p}$ to $E$ and integrate it with the descriptor $\mathbf{d}$ by element-wise addition $\oplus$:
\begin{equation}
    \mathcal{X} = \mathcal{D} \oplus \textbf{MLP}_{\text{enc}}\left(\mathcal{P}\right),
\end{equation}
where $\mathcal{X}\in\mathbb{R}^{E\times N}$ is the matrix of positional-aware descriptors and $\mathcal{D}\in\mathbb{R}^{E\times N}$ is the matrix consisting of original descriptors $\mathbf{d}_i, i\in\{1, ..., N\}$. 
Position embedding is common in sequence-to-sequence learning problems~\cite{gehring2017convolutional, vaswani2017attention}, enhancing the model's ability to capture contextual cues. In our case, it improves the embeddings' distinctiveness by allowing the GNN to distinguish between different embeddings of point clouds at different positions in the subgraph, thus capturing the contextual and viewpoint information more effectively (See~\secref{sec:ablations}). 

\noindent
\textbf{Multi-head Attentional GNN:}
As stated before, our graph between pairs of subgraphs is fully connected and multiplex,~\ie~it comprises two types of edges: intra edges $\mathcal{E}_{\text{intra}}$ which are the edges between the nodes within one subgraph and inter edges $\mathcal{E}_{\text{inter}}$ which are the edges across the subgraphs in the input pair. Borrowing the terminology of \emph{message passing} in GNN from~\cite{battaglia2018relational}, we aggregate the messages carried through edges $e_{ij}\in \mathcal{E}$ to keynode $i$,~\ie~$\mathbf{m}_{e_{i}}:=\{\mathbf{m}_{e_{ij}} |~\forall j: e_{ij}\in\mathcal{E}\}$. The AGNN is composed of multiple layers. At each layer ${\ell}\in\{1,...,L\}$, the message passing update,~\ie~updating the intermediate feature ${}^{(\ell)}\mathbf{x}_i$ of keynode $\mathcal{V}_i$, is concurrently conducted by aggregating messages across all the edges for all the keynodes once for subgraph $\mathcal{S}_A$ and once for $\mathcal{S}_B$ using:
\begin{equation}
    {}^{(\ell+1)}\mathbf{x}_i =
        {}^{(\ell)}\mathbf{x}_i
        + \mathcal{M}_t({}^{(\ell)}\mathbf{x}_i, \mathbf{m}_{e_i}),
\end{equation}
where $\mathcal{M}_t$ is the message function. Following~\cite{sarlin20superglue}, we use a MLP, albeit modfied (See \secref{sec:implementation}), and concatenate messages with intermediate representations,~\ie~$\mathcal{M}_t(\mathbf{x},\mathbf{m}_e) = \textbf{MLP}([\mathbf{x}\|\mathbf{m}_e])$, where $[.\|.]$ denotes concatenation operator. 

Following the attention mechanism described in~\cite{vaswani2017attention}, we aggregate messages received by each keynode through edges within (intra-attention) and across (inter-attention) subgraphs $\mathcal{S}_A$ and $\mathcal{S}_B$. To this end, we consider the receiver keynode to be in subgraph $\mathcal{S}_R$ and the entire sender keynodes to be in $\mathcal{S}_S$ such that $(\mathcal{S}_R,\mathcal{S}_S)\in\{\mathcal{S}_A,\mathcal{S}_B\}^2$.
\begin{equation}
\label{eq:attention}
    \begin{split}
        \mathbf{Q} &= \mathbf{W}_Q\ \mathcal{X}_R + \mathbf{b}_Q,\\
        \begin{bmatrix}\mathbf{K}\\\mathbf{V}\end{bmatrix} &=
        \begin{bmatrix}\mathbf{W}_K\\\mathbf{W}_V\end{bmatrix} \mathcal{X}_S
        + \begin{bmatrix}\mathbf{b}_K\\\mathbf{b}_V\end{bmatrix}, \\
    \end{split}
\end{equation}
where matrices $\mathbf{Q}$, $\mathbf{K}$ and $\mathbf{V}$ consist of query vectors (features of keynodes that are being attended to) in $\mathcal{S}_R$, $\mathbf{K}$ keys (features that are used to compute the attention scores) and $\mathbf{V}$ values (features that are weighted by the attention scores to compute the output) in $\mathcal{S}_S$, respectively. $\mathcal{X}_R$ and $\mathcal{X}_S$ are the matrices packing the intermediate features $\mathbf{x}_i$ from $\mathcal{S}_R$ and $\mathbf{x}_j$ from $\mathcal{S}_S$, respectively. Matrices $\mathbf{W}_Q$, $\mathbf{W}_K$ and $\mathbf{W}_V$ are linear transformation weights computed within training along with biases $\mathbf{b}_Q$, $\mathbf{b}_K$ and $\mathbf{b}_V$. The entire messages can now be computed by scaled-dot product attention  as follows:
\begin{equation}
\label{eq:msgs_aggr}
        \mathbf{m}_{e} =  \text{softmax}\left(\frac{\mathbf Q^\top \mathbf K}{\sqrt{d_k}}\right) \mathbf{V},
\end{equation}
where $d_k$ is the keys dimension. For large $d_k$ values, the scaling is essential to avoid large magnitudes likely to cause minimal gradients when using a softmax function~\cite{vaswani2017attention}. 
We use multi-head attention to capture different types of information and relationships within the input more effectively. 
\vspace{-0.2cm}

\subsection{Classification Layer}
The output of the AGNN block is two tensors representing the final descriptors of subgraph $\mathcal{S}_A$ and $\mathcal{S}_B$, $\mathcal{F}^\text{A} = \{\mathbf{f}_i^{\text{A}}\}_{i=1}^N \in \mathbb{R}^{E \times N}$ and $\mathcal{F}^\text{B} = \{\mathbf{f}_j^{\text{B}}\}_{j=1}^M \in \mathbb{R}^{E \times M}$. The final prediction in~\pgat~is performed using cosine similarity to produce a similarity matrix $\mathbf{S}=\{s_{ij}\} \in \mathbb{R}^{N \times M}$. Element $s_{ij}$ is obtained from: 
\begin{equation}
    s_{ij} = \frac{\langle \mathbf{f}_i^{\text{A}} \; , \; \mathbf{f}_j^{\text{B}} \rangle}{\|\mathbf{f}_i^{\text{A}} \| \| \mathbf{f}_j^{\text{B}} \|}, 
\end{equation}
where $\langle.,.\rangle$ and $\|.\|$ denote the inner product and the L2 norm, respectively.

Based on the definition of PR, the descriptors of two nearby point clouds should be similar,~\ie~$s_{ij}$ close to $1$, while the descriptors of two dissimilar point clouds should be distinct,~\ie~$s_{ij}$ close to $-1$. Since there can be multiple pairs of nearby point clouds on a given pair of two subgraphs, PR based on two subgraphs becomes a multi-label binary classification problem. Additionally, we map the scores in $\mathbf{S}$ with the range $\left[-1, 1\right]$ to $\mathbf{P}=\{p_{ij}\}$ with the range $\left[0, 1\right]$ using:
\begin{align}
    p_{ij} = s_{ij} \times 0.5 + 0.5,
\end{align}
and interpret it as the probability that point cloud $\mathcal{P}_i$ from subgraph $\mathcal{S}_A$ and point cloud $\mathcal{P}_j$ from subgraph $\mathcal{S}_j$ represent the same place. Our classification problem can now be defined as a stochastic optimisation problem measuring the similarity between two probabilities. 

\subsection{Loss}
Reformulating the binary classification problem as stochastic optimisation, we minimise the Kullback-Leibler (KL) divergence $D_{KL}(\mathbf{y}\|\mathbf{P})=-\sum_{i=1}^Np_i \log(\frac{p_i}{y_i})$ between the predicted probabilities $p_i$ and the true probabilities $y_i$ (ground truth). The true probabilities follow the Bernoulli distribution,~\ie~the probability that random variable $x\in\{0,1\}$ belongs to a class ($x=1$) is $p(x=1)=p$, otherwise $p(x=0)=1-p$.  Using the Bernoulli distribution properties, the KL divergence is converted into Binary Cross Entropy (BCE)~\cite{de2005tutorial}.
Since we have multiple separate classifications to perform (between the keynodes in pairs of subgraphs), our final BCE loss is defined as follows:
\begin{equation}
\label{eq:loss}
\resizebox{1.0\hsize}{!}{$
        \mathcal{L}\left(\mathbf{y}, \mathbf{P} \right) = -\sum^N_{i=1} \sum^M_{j=1} \omega_{ij} \left( y_{ij} \cdot \log p_{ij} 
     + \left( 1 - y_{ij}\right) \cdot \log\left( 1 - p_{ij}\right) \right),
    $}
\end{equation}
where $y_{ij}\in\{0,1\}$ is the ground truth label. 
If point cloud $\mathcal{P}_i$ and point cloud $\mathcal{P}_j$ represent the same place $y_{ij}=1$, otherwise, $y_{ij}=0$.~$\omega_{ij}$ is a scalar hyperparameter to indicate whether the point clouds pair $\left(\mathcal{P}_i, \mathcal{P}_j\right)$ contributes in the loss function based on the conditions we follow to select positive and negative pairs (See~\secref{sec:implementation}).

\section{Experiments}
\label{sec:expe}

We briefly introduce the datasets used, followed by evaluation settings and the implementation details. Comparisons are made between our proposed network and three state-of-the-art architectures with different backbones, serving as baselines. Additionally, we assess our network's performance in comparison to existing lidar PR approaches. We then provide detailed ablation studies to verify the network design and the impact of each component on performance gains.
\vspace{-0.2cm}

\subsection{Datasets}
We use three publicly-available large-scale datasets for evaluation. We detail the characteristics of each dataset below.

\noindent
\textbf{Oxford RobotCar dataset~\cite{RobotCarDatasetIJRR}} 
has been widely used for lidar PR, which is a processed subset of the overall Oxford RobotCar dataset~\cite{RobotCarDatasetIJRR}. It consists of point clouds captured by travelling a route ($\sim$10 km) 44 times across Oxford, UK, over a year. To assess performance on the Oxford RobotCar dataset, point clouds from one trip are taken as queries and matched against point clouds from other trips in an iterative process~\cite{uy2018pointnetvlad}. The training and testing dataset split introduced by Uy~\etal~\cite{uy2018pointnetvlad} is followed in this work. In total, $\sim$ 24.7k point clouds were used for training and testing.

\noindent
\textbf{In-house dataset~\cite{uy2018pointnetvlad}} consists of data from three regions in Singapore -- a University Sector (U.S.), a Residential Area (R.A.), and a Business District (B.D.). Similar to the evaluation on the Oxford dataset, we use the standard test dataset split described in~\cite{uy2018pointnetvlad}. Point clouds were collected on a car travelling a path in U.S. ($\sim$ 10 km), R.A. ($\sim$ 8 km) and B.D. ($\sim$ 5 km) 5 times at different times. Point clouds from a single trip are used as queries and evaluated iteratively with point clouds from the remaining trips as database~\cite{uy2018pointnetvlad}. In-house dataset used at the test time only to demonstrate generalisability. In total, $\sim$ 4.5k point clouds were used for testing.

\noindent
\textbf{MulRan dataset~\cite{Kim2020MulRanMR}} consists of traversals of several environments in South Korea -- the Daejeon Convention Center (DCC) (3 runs each $\sim$ 5 km), the Riverside (3 runs each $\sim$ 6 km) of Daejeon city, the Korea Advanced Institute of Science and Technology (KAIST) (3 runs each $\sim$ 7 km) and Sejong city (Sejong) (3 runs each $\sim$ 23 km). 
We use the DCC and Riverside environments, training with DCC sequences 1 and 2 and testing on sequence 3, and training with Riverside sequences 1 and 3 and testing on sequence 2. KAIST three sequences are used as unseen only for evaluation. 
Because the average distance of point clouds in the MulRan dataset is $\sim$ 1 m, we only use the point clouds with a minimum of 20 m apart, resulting in a total of $\sim$ 2.5k point clouds used for training and testing.
\vspace{-0.4cm}

 \subsection{Evaluation Criteria}
The datasets described above include UTM coordinates,~\ie~ground truth obtained from IMU/GPS for Oxford/in-house and IMU/GPS/SLAM for MulRan, for each point cloud. Using this ground truth, we select a 25 m threshold to classify successful place recognition events (retrievals). We compare the cosine similarity between the refined global descriptors of each query in a run with the refined global descriptors of the remaining point clouds covering the same region in the database. 
For comparison, we use AR@N (and its varieties,~\ie~AR@1 and AR@1\%), commonly used for lidar PR performance. This metric measures the percentage of correctly localised queries where at least one of the top-N database predictions matches the query. A perfect AR@N score would be 100\%, meaning all the possible revisits are correctly identified.
\vspace{-0.4cm}
\setlength{\tabcolsep}{5pt}
\begin{table*}[t]
\caption{\small{Across all the baseline and all dataset combinations, our \pgat{} exhibits performance gains in place recognition (measured by Recall@1). The best performance is indicated in bold.}}
\label{tab:comparison}
\large
\resizebox{\textwidth}{!}{\begin{tabular}{clllllllllllllllllll}
\hline
\multicolumn{1}{l}{\multirow{2}{*}{}} & &  & \multicolumn{14}{c}{\textbf{Tested on:}}& &\\
\multicolumn{1}{l}{} & \multicolumn{1}{c}{} & & \multicolumn{2}{c}{Oxford} & \multicolumn{2}{c}{DCC} & \multicolumn{2}{c}{Riverside} & \multicolumn{2}{c}{B.D.} & \multicolumn{2}{c}{R.A.} & \multicolumn{2}{c}{U.S.}& \multicolumn{2}{c}{KAIST} & \multicolumn{2}{c}{Average}\\ 
\multicolumn{1}{l}{\textbf{Trained on:}} & \multicolumn{1}{c}{} & & AR@1 & AR@1\% & AR@1 & AR@1\% & AR@1 & AR@1\% & AR@1 & AR@1\% & AR@1 & AR@1\% & AR@1 & AR@1\% & AR@1 & AR@1\% & AR@1 & AR@1\% \\\hline
\multirow{6}{*}{Oxford} & \multirow{2}{*}{PointNetVLAD} & Baseline & 62.8 & 83.8 & 69.3 & 81.6 & 37.0 & 56.6 & 56.2 & 65.3 & 52.3 & 68.4 & 63.1 & 79.0 & 57.2 & 73.1 & 56.8& 72.5\\
 & & \pgat &  \textbf{96.5} & \textbf{99.9} & \textbf{80.8} & \textbf{92.1} & \textbf{57.0} & \textbf{83.5} & \textbf{89.2} & \textbf{98.6} & \textbf{76.3} & \textbf{99.6} & \textbf{79.0} & \textbf{98.2} & \textbf{73.8} & \textbf{82.3}& \textbf{78.9}& \textbf{93.5}\\ \cline{2-19} 

 & \multirow{2}{*}{MinkLoc3D} & Baseline & 93.0 & 97.9 & 77.7 & 91.2 & 46.3 & 81.1 & 81.5 & 88.5 & 80.4 & 91.2 & 86.7 & 95.0 & 78.8 & 91.6 &77.8& 90.9\\
 & & \pgat & \textbf{98.0} & \textbf{99.9} & \textbf{93.9} & \textbf{94.7} & \textbf{75.6} & \textbf{86.4} & \textbf{98.0} & \textbf{99.8} & \textbf{94.3} & \textbf{100.0} & \textbf{98.0} & \textbf{100.0} & \textbf{89.1} & \textbf{94.2} & \textbf{92.4} & \textbf{96.4} \\ \cline{2-19} 

 & \multirow{2}{*}{PPT-Net} & Baseline & 93.5 & 98.1 & 76.0 & 89.6 & 39.4 & 84.3 & 84.6 & 90.0 & \textbf{84.1} & 93.3 & 90.1 & 97.5 & 77.4 & 88.3 & 77.9 & 91.6\\
\multicolumn{1}{l}{} & & \pgat & \textbf{97.4} & \textbf{99.9} & \textbf{93.6} & \textbf{99.2} & \textbf{73.1} & \textbf{94.0} & \textbf{89.0} & \textbf{98.1} & 82.9 & \textbf{99.8} & \textbf{90.9} & \textbf{99.9} & \textbf{82.3} & \textbf{91.4} & \textbf{87.4}& \textbf{97.5} \\\hline

\multirow{5}{*}{DCC} & \multirow{2}{*}{PointNetVLAD} & Baseline & 40.9 & 57.6 & 79.3 & 90.9 & 52.5 & 81.6 &  52.2 & 60.5 & 45.0 & 58.5 & 49.1 & 65.5 & 64.9 & 78.9 & 54.8& 70.5\\
\multirow{5}{*}{Riverside} &  & \pgat &  \textbf{65.6} & \textbf{93.9} & \textbf{96.9} & \textbf{99.9} & \textbf{83.2} & \textbf{96.5} & \textbf{52.5} & \textbf{76.9} & \textbf{61.9} & \textbf{95.9} & \textbf{58.5} & \textbf{96.4} & \textbf{79.0} & \textbf{91.4} &\textbf{71.1}& \textbf{93.0}\\ \cline{2-19} 

 & \multirow{2}{*}{MinkLoc3D} & Baseline & 68.9 & 81.5 & \textbf{95.4} & \textbf{99.2} & 78.1 & 92.7 & 79.4 & 85.7 & 79.5 & 88.7 & 84.0 & 92.5 & \textbf{85.3} & \textbf{93.5} & 81.5& 90.5\\
 & & \pgat & \textbf{78.9} & \textbf{97.3} & 93.2 & 98.9 & \textbf{87.3} & \textbf{95.6} & \textbf{95.4} & \textbf{99.9} & \textbf{81.3} & \textbf{99.2} & \textbf{89.5} & \textbf{99.9} & 82.2 & 92.8 & \textbf{86.8}& \textbf{97.7}\\ \cline{2-19} 

 & \multirow{2}{*}{PPT-Net} & Baseline & 63.3 & 78.8 & 91.7 & 99.6 & 66.1 & 94.6 & 66.4 & 75.5 & 65.7 & 81.8 & 77.2 & 89.4 & \textbf{83.6} & \textbf{93.6} & 73.4& 87.6\\
\multicolumn{1}{l}{} & & \pgat & \textbf{69.1} & \textbf{96.3} & \textbf{98.9} & \textbf{100.0} & \textbf{87.3} & \textbf{99.4} & \textbf{77.3} & \textbf{94.6} & \textbf{74.9} & \textbf{97.9} & \textbf{80.2} & \textbf{99.8} & 78.8 & 89.8 & \textbf{80.9}& \textbf{96.8}\\ \hline

\end{tabular}}
\end{table*}

\subsection{Implementation Details}
\label{sec:implementation}
We implemented \pgat~in PyTorch, and Adam optimiser was used. The learning rate was set to $1e^{-4}$ %
without learning rate decay. Our model was trained with batch size 256 for 1500 epochs on Oxford and MulRan. The number of attention layers was set to 9 with four heads for multi-head AGNN, resulting in $\sim12$ million parameters (in total) for learning. We used this single configuration across all the experiments. 

We create an adjacency matrix using a travel distance threshold to generate fully connected subgraphs. Using this matrix, we create subgraphs that include (key)nodes, each encompassing both the pose and its corresponding feature information. The adjacency matrix plays a crucial role in the training process for pairing subgraphs, ensuring that nodes within each subgraph are fully interconnected. Upon pairing subgraphs, we merge the nodes and features from both subgraphs, forming a fully connected graph. Nodes define edges in between, and features are populated with descriptors obtained from an existing PR method. We randomly pair subgraphs from the database to select two subgraphs as input. Because most of the pairs of subgraphs do not overlap, the ground truth of the similarity matrix is almost a zero matrix, not allowing the model to update the parameters properly. Therefore, we defined a positive rate and set it to 30\%, forcing the model to have a 30\% probability of having subgraph pairs with at least one pair of positive nodes.

Subgraph generation is based on the robot's travelling distance. A subgraph starts from node $\mathcal{V}_i$ and stops at node $\mathcal{V}_{i + n}$ when the travelling distance exceeds a threshold. The stride of the subgraph generation is one, \ie~the next subgraph starts from node $\mathcal{V}_{i+1}$. Since the number of subgraph nodes can vary, we use a boolean mask to deal with padding nodes. Also, we implemented a customised version of layer normalisation (compatible with the dynamic behaviour of subgraphs) in the MLP blocks, calculating all nodes' mean and standard deviation except padding nodes. This approach effectively handles the dynamic node number within subgraphs.

During testing, because one query node can appear in multiple subgraphs (unlike point cloud pair-wise comparison), we implemented an average scheme in which the similarity scores related to a query node are averaged when the query node is compared against the entire nodes from the database,~\ie~$\overline{s}_{ij}=\frac{1}{C_1C_2}\sum_{n_1=1}^{C_1}\sum_{n_2=1}^{C_2}s_{ij}(\mathbf{d}_i^{(n_1)},\mathbf{d}_j^{(n_2)})$, where $C_1$ and $C_2$ are the maximum numbers of subgraphs from which the $i$-th query node and $j$-th database node are seen, respectively.      
\vspace{-0.35cm}

\begin{table}[t]
\caption{\small{Average recall (\%) at top 1\% (AR@1\%) and top 1 (AR@1) for the state-of-the-art lidar-based PR models trained on the Oxford RobotCar. Our P-GAT (trained on top of MinkLoc3D) performs best on all benchmarks.}}
\label{tab:benchmarks}
\huge
\resizebox{\linewidth}{!}{\begin{tabular}{lccccccccc}
\hline
 \multicolumn{1}{c}{}  & \multicolumn{2}{c}{Oxford} & \multicolumn{2}{c}{U.S.} & \multicolumn{2}{c}{R.A.} & \multicolumn{2}{c}{B.D.}\\ 
\multicolumn{1}{c}{}  & AR@1 & AR@1\% & AR@1 & AR@1\% & AR@1 & AR@1\% & AR@1 & AR@1\%\\\hline
PointNetVLAD~\cite{uy2018pointnetvlad} & 62.8 & 80.3 & 63.2 & 72.6 & 56.1 & 60.3 & 57.2 & 65.3\\
PCAN~\cite{zhang2019pcan}  & 69.1& 83.8& 62.4& 79.1& 56.9& 71.2& 58.1& 66.8\\
LPD-Net~\cite{liu2019lpd}  & 86.3 & 94.9 & 87.0 & 96.0 & 83.1 & 90.5 & 82.5 & 89.1\\
EPC-Net~\cite{hui2022efficient}  & 86.2 & 94.7 & \---- & 96.5 & \---- & 88.6 & \---- & 84.9\\   
HiTPR~\cite{hou2022hitpr}  & 86.6 & 93.7 & 80.9 & 90.2 & 78.2 & 87.2 & 74.3 & 79.8\\
SOE-Net~\cite{xia2021soe}  & 89.4 & 96.4 & 82.5 & 93.2 & 82.9 & 91.5 & 83.3 & 88.5\\
MinkLoc3D~\cite{jacek20minkloc}  & 93.0 & 97.9 & 86.7 & 95.0 & 80.4 & 91.2 & 81.5 & 88.5\\
NDT-Transformer~\cite{zhou2021ndt}  & 93.8 & 97.7 & \---- & \---- & \---- & \---- & \---- & \----\\
PPT-Net~\cite{hui2021pyramid}  & 93.5 & 98.1 & 90.1 & 97.5 & 84.1 & 93.3 & 84.6 & 90.0\\
SVT-Net~\cite{fan2022svt}  & 93.7 & 97.8 & 90.1 & 96.5 & 84.3 & 92.7 & 85.5 & 90.7\\
MinkLoc3D-S~\cite{zywanowski2021minkloc3d}  & 92.8 & 81.7 & 83.1 & 67.7 & 72.6 & 57.1 & 70.4 & 62.2\\
PVT3D~\cite{xia2022pvt3d}  & 95.6 & 98.5 & 92.9 & 97.9 & 89.5 & 94.8 & 87.9 & 92.1\\
\pgat{} (Ours)  & \textbf{98.0} &  \textbf{99.9} & \textbf{98.0} & \textbf{100.0} & \textbf{94.3} & \textbf{100.0} & \textbf{98.0} & \textbf{99.8} \\\hline
\end{tabular}}
\vspace{-0.3cm}
\end{table}

\begin{figure*}[t]
    \centering
    \includegraphics[width=0.24\linewidth]{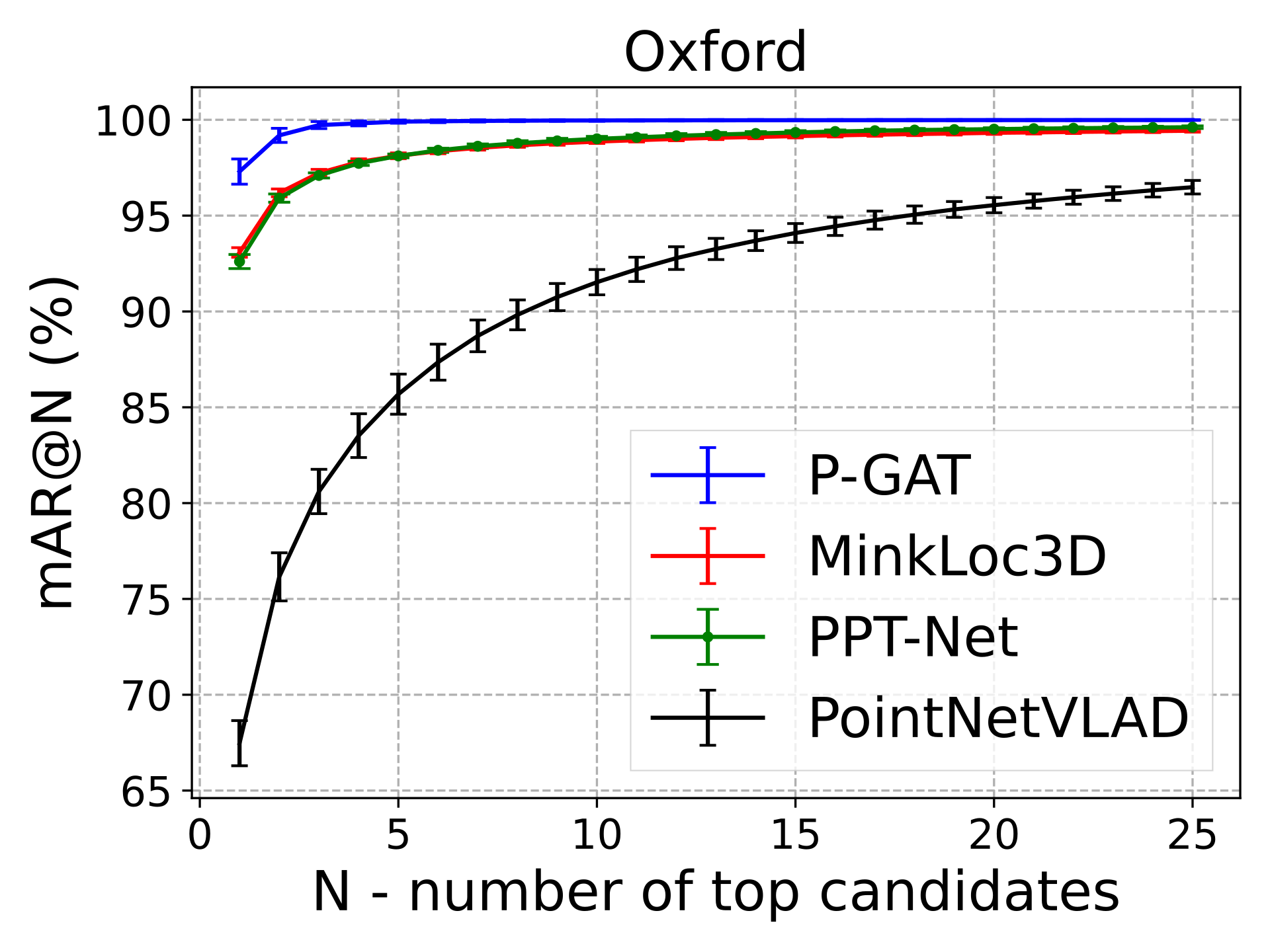}%
    \includegraphics[width=0.24\linewidth]{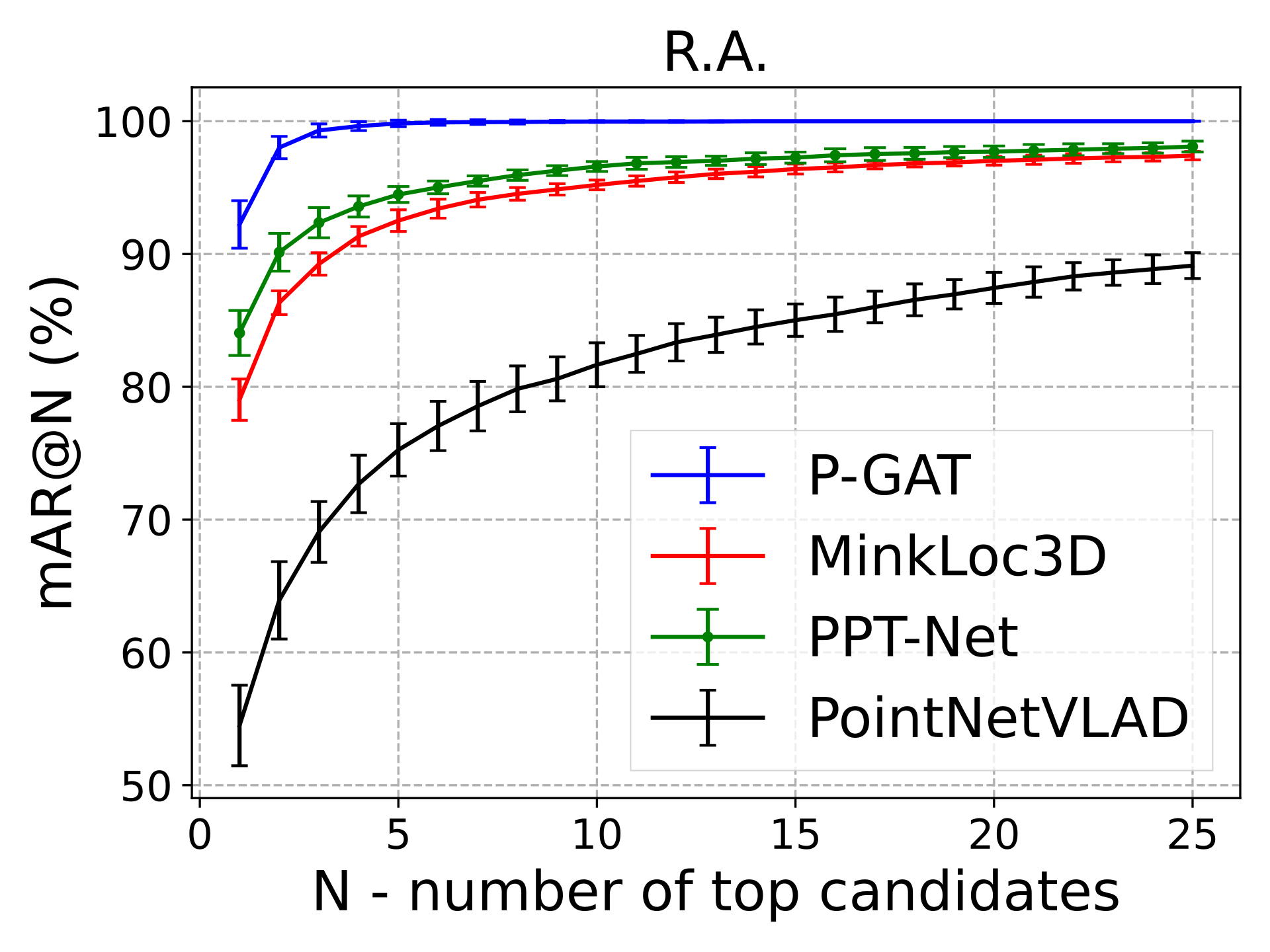}%
    \includegraphics[width=0.24\linewidth]{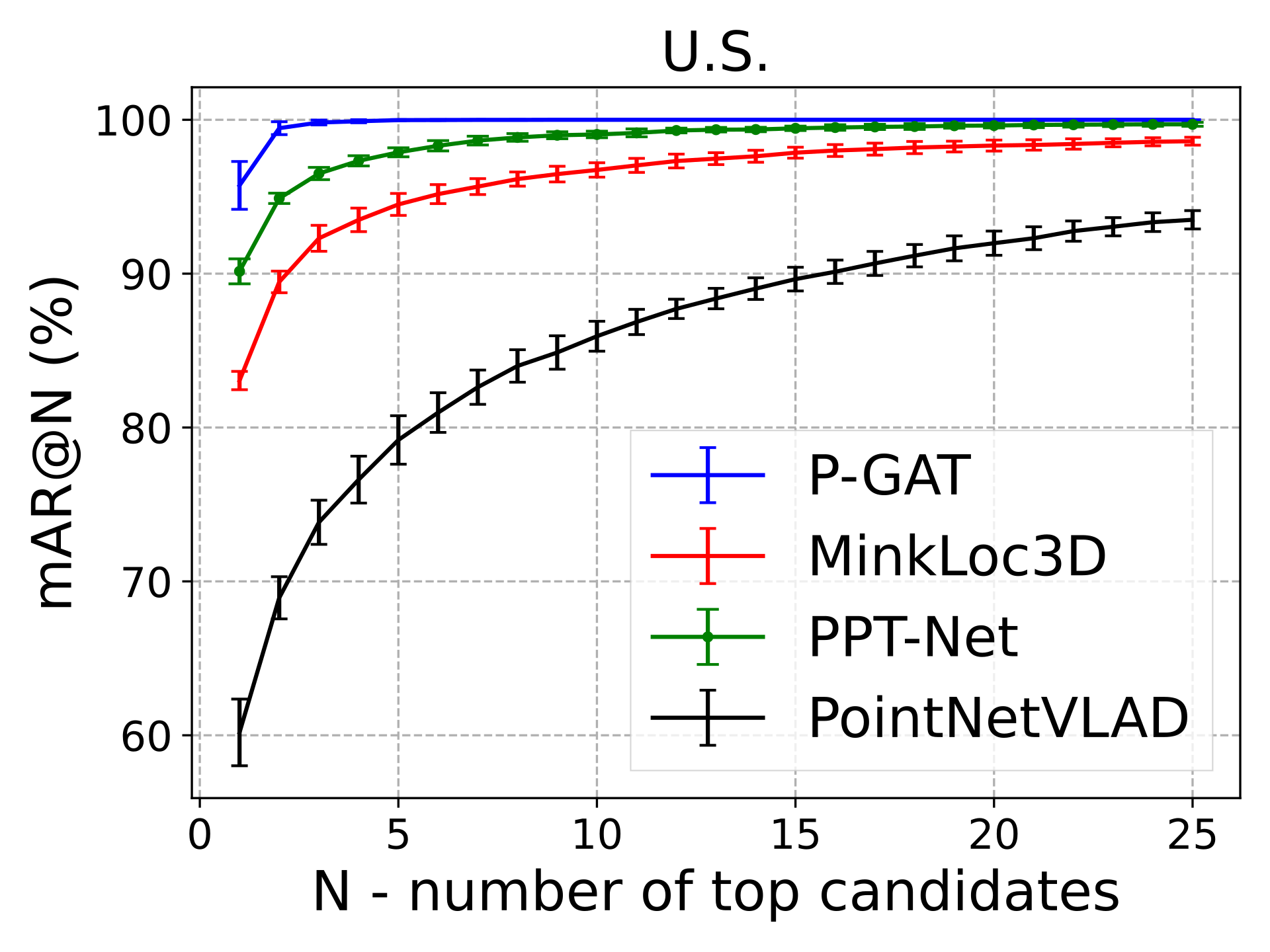}%
    \includegraphics[width=0.24\linewidth]{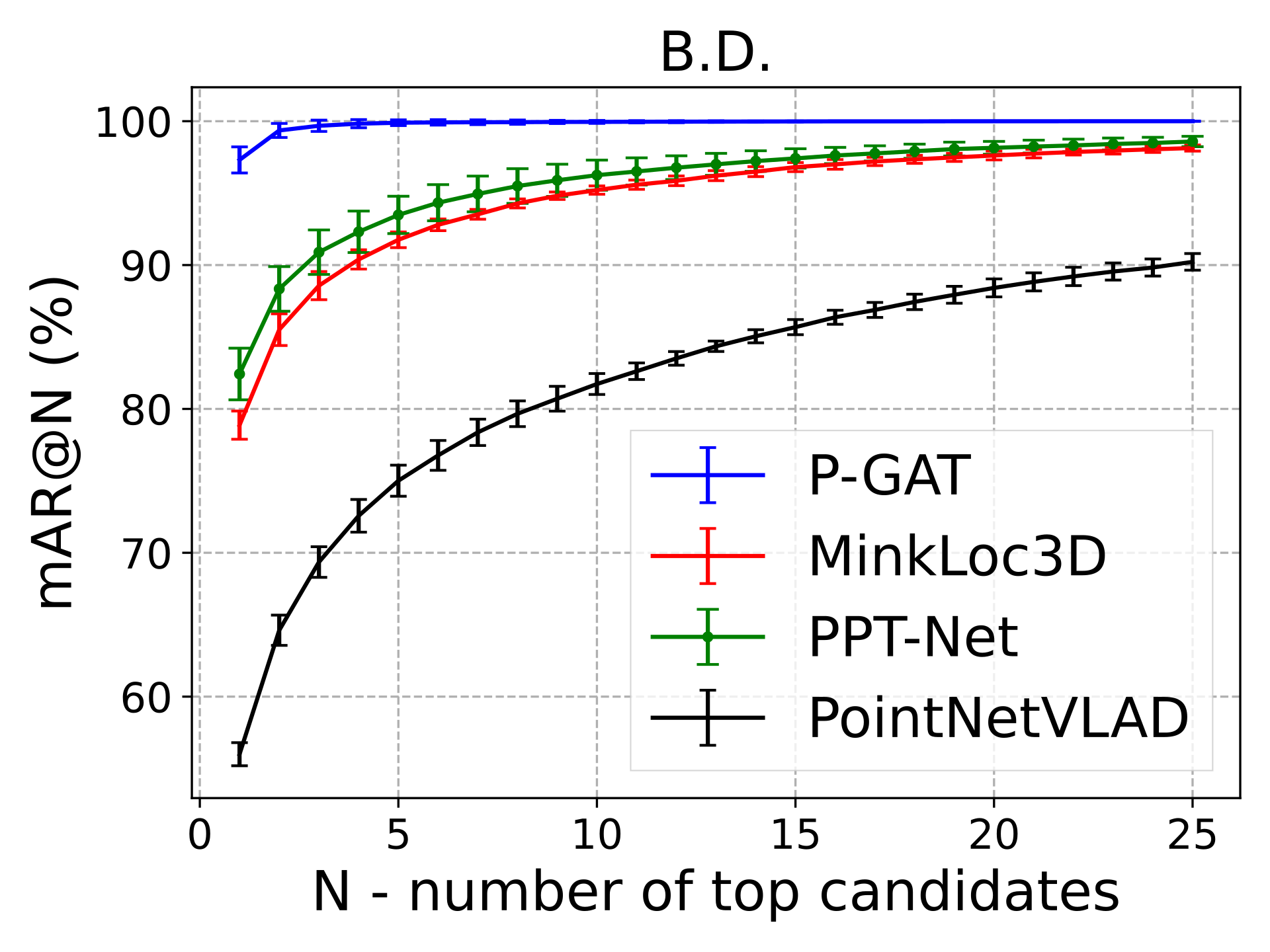}
    \caption{\small{Variance analysis: Following the training and testing split protocol of~\cite{uy2018pointnetvlad}, we re-trained P-GAT, MinkLoc3D, PointNetVLAD and PPT-Net 5 times, with a fixed parameter setting, on the Oxford training set and evaluated on the Oxford/in-house testing split.}}
\label{fig:variance}
\end{figure*}
\subsection{Comparison to State-of-the-Art}
\label{sec:comparison}
To show that~\pgat~is backbone agnostic, we separately trained the model on Oxford RobotCar~\cite{RobotCarDatasetIJRR} and MulRan datasets~\cite{Kim2020MulRanMR} using descriptors $\mathbf{d} \in \mathbb{R}^{256}$ obtained from the backbone models PointNetVLAD~\cite{uy2018pointnetvlad}, MinkLoc3D~\cite{jacek20minkloc}, and PPT-Net~\cite{hui2021pyramid}. Descriptors of these baselines were generated after training the models on the training splits described earlier. 

\tabref{tab:comparison} summarises the average recall@1\% (AR@1\%) and average recall@1 (AR@1) of the results of our \pgat~and three other baseline models. Across all the experiments, \pgat~displays substantial performance gains (on average above 10\%) compared to the baselines, indicating that if \pgat~is integrated with any backbone, it improves the performance due to spatiotemporal information. Additionally, following the trend of models' performance when trained on Oxford and tested on MulRan/in-house or trained on MulRan and tested on Oxford/in-house, \pgat~demonstrates more consistent improvement in performance, showing that \pgat~enhances the generalisation ability of the network when train and test data are from different distributions. 

We further compare our \pgat~with the state-of-the-art methods listed in~\tabref{tab:benchmarks}. For this, P-GAT was trained using descriptors extracted from \minkloc trained on the Oxford training subset. Comparing \pgat~and \minkloc baseline, we observe that AR@1 for Oxford, U.S., R.A., and B.D. improves by 5\%, 11.3\%, 13.9\%, and 16.5\%, respectively. Additionally, although we used \minkloc descriptors, the substantial gain obtained by \pgat~allows us to outperform the other state-of-the-art methods, such as PPT-Net, SVT-Net and PVT3D, which reported higher performance than that of vanilla \minkloc. AR@1\% scores for all datasets are all higher than 99\%. High AR@1\% indicates that \pgat~enables the identification of all possible revisits in a small subset of top candidates (1\% of point clouds in the database).

To evaluate the performance variance of our model compared to the baselines, we trained \pgat, PointNetVLAD, \minkloc and PPT-Net 5 times on the Oxford training subset and tested on the Oxford/in-house testing subset using fixed settings. \figref{fig:variance} shows mean AR@N recalls (average of AR@N over 5 experiments) and their standard deviations for all four models. The variance of our model is comparable with that of \minkloc and PPT-Net, and it decreases by increasing the number of top candidates. PointNet-VLAD shows greater variance than the other methods indicating the network has more nondeterministic behaviour. 
The recall curves provide a visual representation of the superior performance of our \pgat~model, demonstrating the potential of our attention-based approach in improving the accuracy and stability of place recognition. 
\subsection{Ablation Studies}
\label{sec:ablations}

We conducted ablation studies to validate our proposed method and the relative contribution of each component. For this purpose, we trained and evaluated \pgat~utilising the \minkloc model trained on the Oxford dataset. This ensures a fair comparison with the results reported in Table~\ref{tab:benchmarks}.

\begin{table}[]
    \caption{\small{Impact of the positional encoding in the subgraph.}}
    \centering
    \scriptsize		
    \begin{tabular}{lcccc}
    \hline
    \multicolumn{1}{c}{}   & \multicolumn{2}{c}{With position}  & \multicolumn{2}{c}{Without position}\\
                           & AR@1 & AR@1\% & AR@1 & AR@1\% \\\hline
    Oxford                 & 98.0 & 99.9   & 95.8 & 99.8 \\
    R.A.                   & 94.3 & 100.0  & 53.3 & 85.1 \\
    U.S.                   & 98.0 & 100.0  & 58.8 & 89.8 \\
    B.D.                   & 98.0 & 99.8   & 59.5 & 82.2 \\ \hline
    \end{tabular}
    \label{tab:ablate_no_pose}
    \vspace{-0.3cm}
\end{table}

\noindent
\textbf{Effects of Positional Encoding:} 
\tabref{tab:ablate_no_pose} compares the performance of our model with and without positional information. The model without positional information shows a negligible drop (2.2\% in AR@1) in performance on the Oxford testing split. However, the model lacking positional information exhibited a considerably inferior performance on the in-house datasets. As seen, AR@1 and AR@1\% decrease more than 40\% and 10\%, respectively, over R.A, U.S. and B.D.
These results demonstrate that positional information is critical for our model's performance, especially when testing in unseen environments. 
\begin{table}[t]
    \caption{\small{Impact of intra- and inter-attention mechanisms.}}
    \centering
    \Huge
    \resizebox{\linewidth}{!}{\begin{tabular}{cc|cccccccc}
    \hline
                  &            & \multicolumn{2}{c}{Oxford} & \multicolumn{2}{c}{U.S.} & \multicolumn{2}{c}{R.A.} & \multicolumn{2}{c}{B.D.} \\
       Intra-Attn      & Inter-Attn      & AR@1 & AR@1\% & AR@1 & AR@1\% & AR@1 & AR@1\% & AR@1 & AR@1\% \\\hline
        -      & -      & 95.6 & 99.3   & 85.9 & 95.2  & 92.1 & 97.7  & 84.8 & 93.9   \\ 
       - & \checkmark     & 96.4 & 99.7   & 94.7 & 99.9 & 92.1 & 99.9 & 97.6 &  \textbf{99.9}  \\
       \checkmark &  -   & 97.0 & \textbf{99.9}   & 78.7 & 98.4 & 87.8 & 98.8 & 73.1 &  89.5  \\
       \checkmark & \checkmark & \textbf{98.0} & \textbf{99.9}   & 98.0 & \textbf{100.0}  & \textbf{94.3} & \textbf{100.0}  & \textbf{98.0} & \textbf{99.8}   \\\hline
    \end{tabular}}
    \label{tab:ablate_attention}
    \vspace{-0.4cm}
\end{table}

\noindent
\textbf{Intra- and Inter-Attention Mechanisms:} 
In this ablation study, we examine the impact of inter- and intra-attention mechanisms on our model's performance.~Table~\ref{tab:ablate_attention} displays the results, indicating that disabling intra-attention produces a slight decrease in performance across all datasets. On the other hand, disabling inter-attention results in a slight improvement in the AR@1 and AR@1\% scores on the Oxford dataset but a significant decrease in AR@1 scores on the in-house datasets. The U.S. and R.A. datasets show a decrease of around 1\% in AR@1\% scores, whereas B.D. experiences a decrease of 10\%. Notably, the model without the full attention mechanism demonstrates the lowest AR@1 and AR@1\% scores on the Oxford dataset, while its performance on the in-house datasets is comparable and sometimes even better than the model without inter-attention. These findings suggest that both inter- and intra- attention mechanisms enhance place recognition accuracy. Moreover, disabling inter-attention may lead to overfitting on the training dataset, resulting in poor generalisation at test-time with distribution shifts.
\begin{table}[t]
    \caption{\small{Impact of the travel distance in the subgraphs.}}
    \centering
    \Huge
    \resizebox{\linewidth}{!}{
    \begin{tabular}{ccccccccccccc}
    \hline
       Distance & \multicolumn{3}{c}{Oxford} & \multicolumn{3}{c}{U.S.} & \multicolumn{3}{c}{R.A.} & \multicolumn{3}{c}{B.D.} \\
        m    &avg.N & AR@1 & AR@1\% &avg.N & AR@1 & AR@1\% &avg.N & AR@1 & AR@1\% &avg.N & AR@1 & AR@1\% \\\hline
        50    &3 & 74.3 & 99.2 & 2 & 57.1 & 89.3 & 2 & 52.3 & 87.6 & 2& 42.1 &  73.8  \\
        100    &5 & 91.1 &  99.9 & 4 & 78.7 & 98.4 & 4 & 87.8 & 98.8 & 4& 73.1 &  89.5  \\
        200    &10 & \textbf{98.0} & \textbf{99.9}  & 8 & \textbf{98.0} & \textbf{100.0} & 8 & \textbf{94.3} & \textbf{100.0} & 8 & \textbf{98.0} & \textbf{99.8}   \\
        300   &15 & 88.6 & 97.1 & 12 & 83.1 & 99.3 & 12 & 69.7 & 98.8 & 12& 90.4 &  99.6  \\ \hline
    \end{tabular}
    }
    \tiny
    \begin{tablenotes}
    \item[*] *avg.N denotes the averaged number of nodes (rounded to nearest integer) in the subgraphs 
    \end{tablenotes}
    \label{tab:ablate_distance}
    \vspace{-0.3cm}
\end{table}

\noindent
\textbf{Travel Distance in Subgraphs:} We conducted an ablation study to investigate the impact of the robot's travelling distance within subgraphs on the performance of our model. Table~\ref{tab:ablate_distance} tabulates the performance results for varying travelling distances (50 m, 100 m, 200 m, and 300 m) in subgraphs. The model trained on subgraphs with a 50 m length exhibits the lowest AR@1 and AR@1\% scores on all datasets. 
As travelling distances in subgraphs increase, the model performance also improves. The optimal results were obtained from the subgraphs with a travelling distance of 200 m. However, the model trained on subgraphs with a travelling distance of 300 m exhibits a slight performance drop. This ablation study demonstrates the importance of the subgraph length in aggregating information within and across subgraphs. Short subgraphs avoid proper receptive fields between similar and dissimilar point clouds. On the other hand, large subgraphs contain irrelevant information, biasing the model performance. 
\vspace{-0.7cm}

\subsection{Runtime Analysis and Memory Usage}
We evaluated the computation time taken on average for each keynode in a subgraph and memory consumption to demonstrate that our presented system can run online. The timing results are collected by running the pre-trained models on a single NVIDIA RTX A3000 Mobile GPU with an Intel(R) Xeon(R) W-11855M CPU @ 3.20GHz CPU.
~\tabref{tab:runtime} reports \pgat's runtime per frame using embeddings extracted by the baselines listed.
Overall, P-GAT adds a constant memory of $\sim1.1$ GiB and $\sim20$ ms to the inference time, demonstrating that, in an end-to-end fashion, the total computation time allows online operation.
\vspace{-0.2cm}
\begin{table}[bt]
\centering
    \caption{\small{Comparison of inference speed and memory consumption.}}
    \tiny	
    \resizebox{\linewidth}{!}{
    \label{tab:runtime}
    \begin{tabular}{lcc}
        \hline

        \multirow{1}{*}{\textbf{Model}}& Memory Usage& Runtime \\\hline
        PointNetVLAD + P-GAT & (2.95+1.13=4.08)GiB     & (22+18=40)ms  \\
        LPD-Net + P-GAT      & (1.94+1.13=3.07)GiB     & (35+18=53)ms  \\
        MinkLoc3D + P-GAT    & (0.85+1.13=1.98)GiB     & (29+18=47)ms  \\
        PPT-Net + P-GAT      & (0.96+1.13=2.09)GiB  & (30+18=48)ms  \\
        \hline
    \end{tabular}
    }
    \vspace{-0.3cm}
\end{table}

\section{Conclusion}
\label{sec:conclusion}
We proposed \pgat~for large-scale place recognition tasks. Through the attention mechanism and graph neural network, we increase the descriptors' distinctiveness by relating the entire point clouds collected in a sequence and in nearby locations in revisit areas. This design allows for exploiting spatiotemporal information. Extensive experiments on the
Oxford dataset, the in-house dataset and the MulRan dataset demonstrate
the effectiveness of the proposed method and its superiority compared to the state-of-the-art. Our proposed \pgat's average performance across key benchmarks is superior by above 10\% over the original baselines demonstrating that \pgat~can be incorporated into any global descriptors, substantially improving their robustness and generalisation ability. 
In future work, we would like to extend the \pgat~for point cloud registration task to accurately determine the position and orientation of retrieved places. 
\vspace{-0.2cm}

\section{Appendix}
In addition to the quantitative results discussed earlier, in the following, we present qualitative examples of P-GAT in action, and visualisations and analysis of the learned attention patterns.

\begin{figure*}[t]
    \centering
    \includegraphics[width=1.0\linewidth]{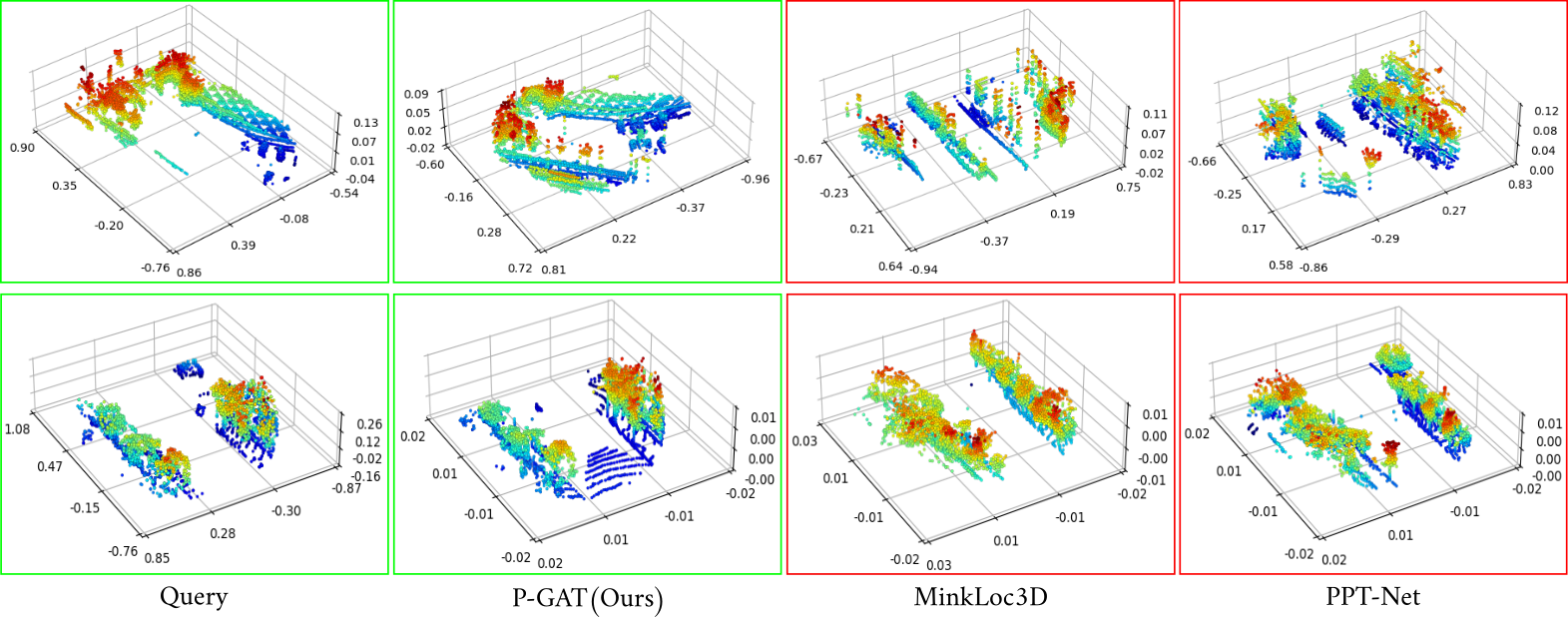}
    \caption{\small{Retrieval examples of our model along with two baseline models (MinkLoc3D and PPT-Net) on R.A. (upper) and Riverside (lower) datasets. True-positive place recognition based on these point clouds is challenging due to the lack of distinct features. However,~P-GAT can recognise similar places correctly, leveraging the spatiotemporal information between neighbour point clouds. The relative distance between the query point cloud and the top candidate selected by P-GAT, MinkLoc3D and PPT-Net is 3.16 m, 801.38 m and 1027.96 m, respectively, for the R.A. exmaple, and 2.83 m, 114.62 m and 127.64 m, respectively, for the Riverside example.}}
\label{fig:example}
\end{figure*}

\begin{figure*}[t]
    \centering
    \includegraphics[width=1.0\linewidth, trim={5cm 0 5cm 0},clip]{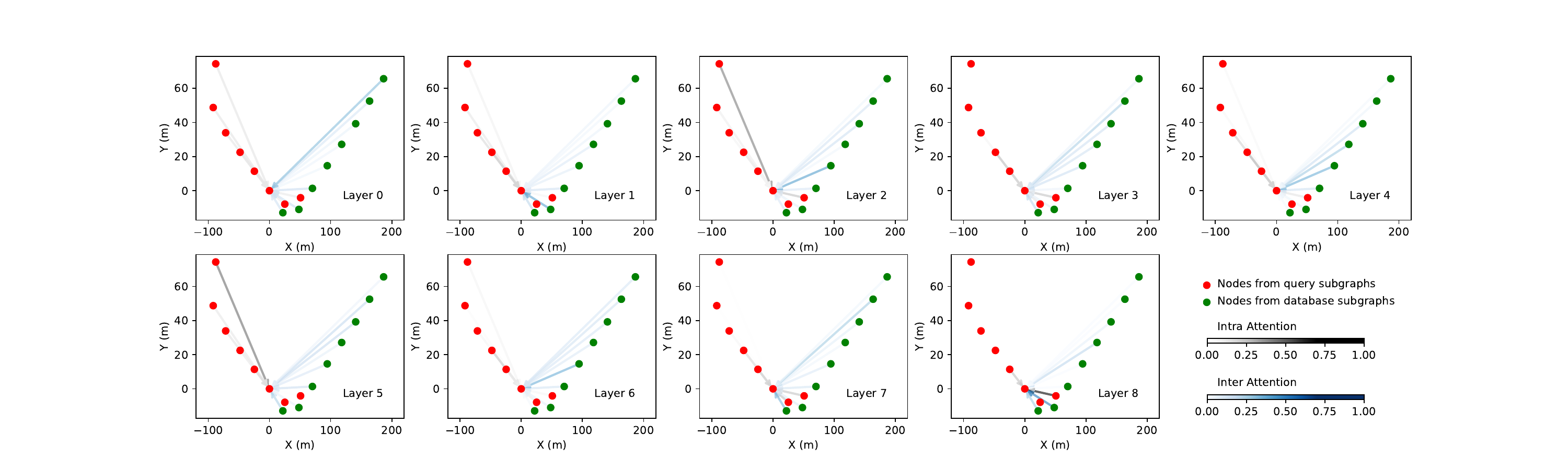}
    \caption{\small{Intra- and Inter- attention pattern, in various layers, between a pair of subgraphs intersecting at a revisit area. Intra edges are colour coded by grey shades, whereas Inter edges by blue shades.}}
\label{fig:attention}
\end{figure*}

\subsection{Qualitative Analysis on Retrieval}
 \figref{fig:example}, visualises some semantically poor point clouds from R.A. (top) and Riverside (bottom) that P-GAT successfully recognised, whereas vanilla MinkLoc3D~\cite{jacek20minkloc} and PPT-Net~\cite{hui2021pyramid} failed in recognition. 
We show the query point clouds on the leftmost column and the top-1 retrieved point clouds from the three models on the following three columns. As seen, \pgat~correctly recognised the place and retrieved the most relevant point cloud despite poor features evidencing the benefit of the aggregation of spatiotemporal information. The third and fourth columns show the retrieval results from MinkLoc3D and PPT-Net. The comparison between our model and the baselines demonstrates the superiority of our model in challenging scenarios.

\subsection{Qualitative Analysis of Intra- and Inter-mechanism}
\figref{fig:attention} demonstrates the attention pattern between two subgraphs (keynodes are indicated by red and green dots) by focusing on a keynode (receiver node) from the query subgraph. As seen in attention layer 0, \pgat~initially attends to all keynodes in the query and dataset subgraphs. By increasing the attention layers, \pgat~focuses on keynodes located nearby the receiver keynode.  Intra-attention aggregates contextual information (in our case spatiotemporal) between point clouds captured consecutively, while inter-attention aggregates information between non-consecutive point clouds in revisited areas, increasing the distinctiveness of features and their invariance to viewpoints and temporal changes. Intermediate layers exhibit oscillating attention spans reflecting the complexity of the learned behaviour.

\subsection{Average Scheme Implementation}
As described in the paper, one keynode can be seen from multiple subgraphs; therefore, more than a similarity score will be computed. This is only the case in inference when ground truth is unavailable. On the other hand, subgraphs in \pgat~are not fixed in length and can have varying numbers of keynodes. Hence we need to deal with this dynamic behaviour of subgraphs as well. To this end, we developed an average scheme to calculate the similarities between query and database keynodes.

\begin{algorithm}[t]
    \caption{Average scheme algorithm} \label{alg:ave}
    \KwInput{subgraph pairs $\{\mathcal{S}_Q, \mathcal{S}_D^l\}$ \break
             \scriptsize{\tcp{$\mathcal{S}_Q$ and $\mathcal{S}_D^l$ are query and $l$-th database subgraphs, $l\in\{1, \dots, L\}$}}}
    \KwOutput{updated similarity matrix $\overline{\mathbf{S}} \in \mathbb{R}^{M \times N}$ \break
    \scriptsize{\tcp{$M$ is the number of keynodes in $\mathcal{S}_Q$, and $N$ is the total number of keynodes (counting duplicates only once) in $\{\mathcal{S}_D^l\}_{l=1}^L$}}}

    Initialise similarity matrix $\overline{\mathbf{S}}$ and counting matrix $\mathbf{C} \in \mathbb{R}^{M \times N}$ with zeros

    \For{$l = 1 : L$}
        {
            $\mathbf{S} =$  \text{\pgat}$( \mathcal{S}_Q, \mathcal{S}_D^l )  $

            \For{$s_{ij}$ in $\mathbf{S}$} 
                {   
                    {\scriptsize{\tcp{$i\in\{1, \dots, M\}, j\in\{1, \dots, N_l\}$, \break$N_l$ is number of keynodes in $\mathcal{S}_D^l$}}}

                    $d$ = $j$ + $l$
                    
                    $\overline{s}_{id} = \overline{s}_{id} + s_{ij}$

                    $c_{id} = c_{id} + 1$

                    {\scriptsize{\tcp{$d\in\{1, \dots, N\}$}}}
                }
        }

    $\overline{\mathbf{S}} = \overline{\mathbf{S}} \oslash \mathbf{C} $ \break \scriptsize{\tcp{$\oslash$ denotes element-wise division}}
\end{algorithm}

\algref{alg:ave} shows the pseudo-code of our average scheme. Our algorithm initialises two matrices, the similarity matrix $\overline{\mathbf{S}} \in \mathbb{R}^{M \times N}$ and the counting matrix $\mathbf{C} \in \mathbb{R}^{M \times N}$, with zeros matrices. We use the proposed \pgat~to calculate the similarity scores in matrix $\mathbf{S} \in \mathbb{R}^{M \times N_l}$ for all pairs of keynodes within the given query and database subgraphs $( \mathcal{S}_Q, \mathcal{S}_D^l )$. We add the similarity score $s_{ij}$ of two keynodes to the corresponding elements $\overline{s}_{id}$ in the similarity matrix and record the number of added similarity scores for $\overline{s}_{id}$ in the corresponding element $c_{id}$ in the counting matrix.~$d$ is the global index of the $j$-th keynode in $\mathcal{S}_D^l$. We map the local keynode index $j$ to the global keynode index $d$ by adding the subgraph's index $l$, as we generated subgraphs using a stride of 1 and considering the incremental indexing of nodes in subgraphs in the database. Finally, we perform an element-wise division between the similarity matrix and the counting matrix to obtain a similarity matrix with elements corresponding to probabilities of whether pairs of keynodes represent the same place. The space complexity of \algref{alg:ave} is $\mathcal{O}(MN), M \ll N$, which is memory efficient. Once the similarity matrix is generated using \algref{alg:ave}, we follow the approach of other place recognition models and select the top K candidates based on their similarity scores.

\section*{Acknowledgements}
This work was funded by CSIRO's Machine Learning and Artificial Intelligence Future Science Platform (MLAI FSP). P.P. and P.M. share senior authorship.
\vspace{-0.2cm}
\balance{}

\bibliographystyle{IEEEtran}
\bibliography{main}

\begin{thebibliography}{10}
\providecommand{\url}[1]{#1}
\csname url@samestyle\endcsname
\providecommand{\newblock}{\relax}
\providecommand{\bibinfo}[2]{#2}
\providecommand{\BIBentrySTDinterwordspacing}{\spaceskip=0pt\relax}
\providecommand{\BIBentryALTinterwordstretchfactor}{4}
\providecommand{\BIBentryALTinterwordspacing}{\spaceskip=\fontdimen2\font plus
\BIBentryALTinterwordstretchfactor\fontdimen3\font minus
  \fontdimen4\font\relax}
\providecommand{\BIBforeignlanguage}[2]{{%
\expandafter\ifx\csname l@#1\endcsname\relax
\typeout{** WARNING: IEEEtran.bst: No hyphenation pattern has been}%
\typeout{** loaded for the language `#1'. Using the pattern for}%
\typeout{** the default language instead.}%
\else
\language=\csname l@#1\endcsname
\fi
#2}}
\providecommand{\BIBdecl}{\relax}
\BIBdecl

\bibitem{huang2023fael}
J.~Huang, B.~Zhou, Z.~Fan, Y.~Zhu, Y.~Jie, L.~Li, and H.~Cheng, ``{FAEL: Fast
  Autonomous Exploration for Large-Scale Environments with a Mobile Robot},''
  \emph{IEEE Robotics and Automation Letters}, 2023.

\bibitem{sarlin2022lamar}
P.-E. Sarlin, M.~Dusmanu, J.~L. Sch{\"o}nberger, P.~Speciale, L.~Gruber,
  V.~Larsson, O.~Miksik, and M.~Pollefeys, ``{LaMAR}: Benchmarking localization
  and mapping for augmented reality,'' in \emph{European Conference on Computer
  Vision}, 2022, pp. 686--704.

\bibitem{knights2023wildplaces}
J.~Knights, K.~Vidanapathirana, M.~Ramezani, S.~Sridharan, C.~Fookes, and
  P.~Moghadam, ``{Wild-Places: A Large-Scale Dataset for Lidar Place
  Recognition in Unstructured Natural Environments},'' in \emph{2023 IEEE
  International Conference on Robotics and Automation (ICRA)}, 2023, pp.
  11\,322--11\,328.

\bibitem{ramezani2023deep}
M.~Ramezani, E.~Griffiths, M.~Haghighat, A.~Pitt, and P.~Moghadam, ``{Deep
  Robust Multi-Robot Re-localisation in Natural Environments},'' in \emph{2023
  IEEE/RSJ International Conference on Intelligent Robots and Systems (IROS)},
  2023.

\bibitem{wang2022transvpr}
R.~Wang, Y.~Shen, W.~Zuo, S.~Zhou, and N.~Zheng, ``{TransVPR: Transformer-Based
  Place Recognition with Multi-Level Attention Aggregation},'' in
  \emph{Proceedings of the IEEE/CVF Conference on Computer Vision and Pattern
  Recognition}, 2022, pp. 13\,648--13\,657.

\bibitem{ali2023mixvpr}
A.~Ali-bey, B.~Chaib-draa, and P.~Gigu{\`e}re, ``{MixVPR: Feature Mixing for
  Visual Place Recognition},'' in \emph{Proceedings of the IEEE/CVF Winter
  Conference on Applications of Computer Vision}, 2023, pp. 2998--3007.

\bibitem{zhang2023etr}
H.~Zhang, X.~Chen, H.~Jing, Y.~Zheng, Y.~Wu, and C.~Jin, ``{ETR: An Efficient
  Transformer for Re-Ranking in Visual Place Recognition},'' in
  \emph{Proceedings of the IEEE/CVF Winter Conference on Applications of
  Computer Vision}, 2023, pp. 5665--5674.

\bibitem{li2023hot}
Z.~Li, C.~D.~W. Lee, B.~X.~L. Tung, Z.~Huang, D.~Rus, and M.~H. Ang,
  ``{Hot-NetVLAD: Learning Discriminatory Key Points for Visual Place
  Recognition},'' \emph{IEEE Robotics and Automation Letters}, 2023.

\bibitem{uy2018pointnetvlad}
M.~A. Uy and G.~H. Lee, ``{PointNetVLAD: Deep Point Cloud Based Retrieval for
  Large-Scale Place Recognition},'' in \emph{The IEEE Conference on Computer
  Vision and Pattern Recognition (CVPR)}, 2018.

\bibitem{jacek20minkloc}
J.~Komorowski, ``{MinkLoc3D: Point Cloud Based Large-Scale Place
  Recognition},'' in \emph{2021 IEEE Winter Conference on Applications of
  Computer Vision (WACV)}, 2021, pp. 1789--1798.

\bibitem{zhang2019pcan}
{Zhang, Wenxiao and Xiao, Chunxia}, ``{PCAN: 3D Attention Map Learning Using
  Contextual Information for Point Cloud Based Retrieval},'' in
  \emph{Proceedings of the IEEE/CVF Conference on Computer Vision and Pattern
  Recognition}, 2019, pp. 12\,436--12\,445.

\bibitem{xia2021soe}
{Xia, Yan and Xu, Yusheng and Li, Shuang and Wang, Rui and Du, Juan and
  Cremers, Daniel and Stilla, Uwe}, ``{SOE-Net: A Self-Attention and
  Orientation Encoding Network for Point Cloud based Place Recognition},'' in
  \emph{Proceedings of the IEEE/CVF Conference on Computer Vision and Pattern
  Recognition}, 2021, pp. 11\,348--11\,357.

\bibitem{hui2021pyramid}
{Hui, Le and Yang, Hang and Cheng, Mingmei and Xie, Jin and Yang, Jian},
  ``{Pyramid Point Cloud Transformer for Large-Scale Place Recognition},'' in
  \emph{Proceedings of the IEEE/CVF International Conference on Computer
  Vision}, 2021, pp. 6098--6107.

\bibitem{ma2022seqot}
J.~Ma, X.~Chen, J.~Xu, and G.~Xiong, ``{SeqOT: A Spatial-Temporal Transformer
  Network for Place Recognition Using Sequential LiDAR Data},'' \emph{IEEE
  Transactions on Industrial Electronics}, 2022.

\bibitem{vidanapathirana2021locus}
K.~Vidanapathirana, P.~Moghadam, B.~Harwood, M.~Zhao, S.~Sridharan, and
  C.~Fookes, ``{Locus: LiDAR-based Place Recognition using Spatiotemporal
  Higher-Order Pooling},'' in \emph{2021 IEEE International Conference on
  Robotics and Automation (ICRA)}, 2021, pp. 5075--5081.

\bibitem{rusu2008aligning}
R.~B. Rusu, N.~Blodow, Z.~C. Marton, and M.~Beetz, ``{Aligning Point Cloud
  Views Using Persistent Feature Histograms},'' in \emph{2008 IEEE/RSJ
  International Conference on Intelligent Robots and Systems}, 2008, pp.
  3384--3391.

\bibitem{rusu2009fast}
R.~B. Rusu, N.~Blodow, and M.~Beetz, ``{Fast Point Feature Histograms (FPFH)
  for 3D Registration},'' in \emph{2009 IEEE International Conference on
  Robotics and Automation}, 2009, pp. 3212--3217.

\bibitem{cop2018delight}
K.~P. Cop, P.~V. Borges, and R.~Dub{\'e}, ``{DELIGHT: An Efficient Descriptor
  for Global Localisation using LiDAR Intensities},'' in \emph{2018 IEEE
  International Conference on Robotics and Automation (ICRA)}.\hskip 1em plus
  0.5em minus 0.4em\relax IEEE, 2018, pp. 3653--3660.

\bibitem{kim2018scan}
G.~Kim and A.~Kim, ``{Scan Context: Egocentric Spatial Descriptor for Place
  Recognition within 3D Point Cloud Map},'' in \emph{2018 IEEE/RSJ
  International Conference on Intelligent Robots and Systems (IROS)}.\hskip 1em
  plus 0.5em minus 0.4em\relax IEEE, 2018, pp. 4802--4809.

\bibitem{dube2017segmatch}
R.~Dub{\'e}, D.~Dugas, E.~Stumm, J.~Nieto, R.~Siegwart, and C.~Cadena,
  ``{SegMatch: Segment Based Place Recognition in 3D Point Clouds},'' in
  \emph{2017 IEEE International Conference on Robotics and Automation (ICRA)},
  2017, pp. 5266--5272.

\bibitem{salti2014shot}
S.~Salti, F.~Tombari, and L.~Di~Stefano, ``{SHOT: Unique signatures of
  histograms for surface and texture description},'' \emph{Computer Vision and
  Image Understanding}, vol. 125, pp. 251--264, 2014.

\bibitem{dube2018segmap}
{Dub{\'e}, Renaud and Cramariuc, Andrei and Dugas, Daniel and Nieto, Juan and
  Siegwart, Roland and Cadena, Cesar}, ``{SegMap: 3D Segment Mapping using
  Data-Driven Descriptors},'' \emph{Robotics: Science and Systems Online
  Proceedings}, vol.~14, 2018.

\bibitem{tinchev2019learning}
{Tinchev, Georgi and Penate-Sanchez, Adrian and Fallon, Maurice}, ``{Learning
  to See the Wood for the Trees: Deep Laser Localization in Urban and Natural
  Environments on a CPU},'' \emph{IEEE Robotics and Automation Letters},
  vol.~4, no.~2, pp. 1327--1334, 2019.

\bibitem{dube2020segmap}
{Renaud Dubé and Andrei Cramariuc and Daniel Dugas and Hannes Sommer and
  Marcin Dymczyk and Juan Nieto and Roland Siegwart and Cesar Cadena},
  ``{SegMap: Segment-based mapping and localization using data-driven
  descriptors},'' \emph{The International Journal of Robotics Research},
  vol.~39, no. 2-3, pp. 339--355, 2020.

\bibitem{ramezani2020online}
{Ramezani, Milad and Tinchev, Georgi and Iuganov, Egor and Fallon, Maurice},
  ``{Online LiDAR-SLAM for Legged Robots with Robust Registration and
  Deep-Learned Loop Closure},'' in \emph{2020 IEEE International Conference on
  Robotics and Automation (ICRA)}.\hskip 1em plus 0.5em minus 0.4em\relax IEEE,
  2020, pp. 4158--4164.

\bibitem{zywanowski2021minkloc3d}
K.~{\.Z}ywanowski, A.~Banaszczyk, M.~R. Nowicki, and J.~Komorowski,
  ``{MinkLoc3D-SI: 3D LiDAR Place Recognition With Sparse Convolutions,
  Spherical Coordinates, and Intensity},'' \emph{IEEE Robotics and Automation
  Letters}, vol.~7, no.~2, pp. 1079--1086, 2021.

\bibitem{komorowski2021egonn}
J.~Komorowski, M.~Wysoczanska, and T.~Trzcinski, ``{EgoNN: Egocentric Neural
  Network for Point Cloud Based 6DoF Relocalization at the City Scale},''
  \emph{IEEE Robotics and Automation Letters}, vol.~7, no.~2, pp. 722--729,
  2021.

\bibitem{lin2017feature}
T.-Y. Lin, P.~Doll{\'a}r, R.~Girshick, K.~He, B.~Hariharan, and S.~Belongie,
  ``{Feature Pyramid Networks for Object Detection},'' in \emph{Proceedings of
  the IEEE Conference on Computer Vision and Pattern Recognition}, 2017, pp.
  2117--2125.

\bibitem{vidanapathirana2022logg3d}
{Vidanapathirana, Kavisha and Ramezani, Milad and Moghadam, Peyman and
  Sridharan, Sridha and Fookes, Clinton}, ``{LoGG3D-Net: Locally Guided Global
  Descriptor Learning for 3D Place Recognition},'' in \emph{2022 International
  Conference on Robotics and Automation (ICRA)}.\hskip 1em plus 0.5em minus
  0.4em\relax IEEE, 2022, pp. 2215--2221.

\bibitem{qi2017pointnet}
{Qi, Charles R and Su, Hao and Mo, Kaichun and Guibas, Leonidas J},
  ``{PointNet: Deep Learning on Point Sets for 3D Classification and
  Segmentation},'' in \emph{Proceedings of the IEEE Conference on Computer
  Vision and Pattern Recognition}, 2017, pp. 652--660.

\bibitem{arandjelovic2016netvlad}
R.~Arandjelovic, P.~Gronat, A.~Torii, T.~Pajdla, and J.~Sivic, ``{NetVLAD: CNN
  architecture for weakly supervised place recognition},'' in \emph{Proceedings
  of the IEEE conference on computer vision and pattern recognition}, 2016, pp.
  5297--5307.

\bibitem{liu2019lpd}
{Liu, Zhe and Zhou, Shunbo and Suo, Chuanzhe and Yin, Peng and Chen, Wen and
  Wang, Hesheng and Li, Haoang and Liu, Yun-Hui}, ``{LPD-Net: 3D Point Cloud
  Learning for Large-Scale Place Recognition and Environment Analysis},'' in
  \emph{Proceedings of the IEEE/CVF International Conference on Computer
  Vision}, 2019, pp. 2831--2840.

\bibitem{qi2017pointnet++}
C.~R. Qi, L.~Yi, H.~Su, and L.~J. Guibas, ``{PointNet++: Deep Hierarchical
  Feature Learning on Point Sets in a Metric Space},'' \emph{Advances in Neural
  Information Processing Systems}, vol.~30, 2017.

\bibitem{vaswani2017attention}
A.~Vaswani, N.~Shazeer, N.~Parmar, J.~Uszkoreit, L.~Jones, A.~N. Gomez,
  {\L}.~Kaiser, and I.~Polosukhin, ``{Attention Is All You Need},''
  \emph{Advances in Neural Information Processing Systems}, vol.~30, 2017.

\bibitem{sarlin20superglue}
P.-E. Sarlin, D.~DeTone, T.~Malisiewicz, and A.~Rabinovich, ``{SuperGlue:
  Learning Feature Matching with Graph Neural Networks},'' in \emph{CVPR},
  2020.

\bibitem{min2021attentional}
C.~Min, J.~Xu, L.~Xiao, D.~Zhao, Y.~Nie, and B.~Dai, ``{Attentional Graph
  Neural Network for Parking-Slot Detection},'' \emph{IEEE Robotics and
  Automation Letters}, vol.~6, no.~2, pp. 3445--3450, 2021.

\bibitem{wang2019attentional}
S.~Wang, Z.~Chen, D.~Li, Z.~Li, L.-A. Tang, J.~Ni, J.~Rhee, H.~Chen, and P.~S.
  Yu, ``{Attentional Heterogeneous Graph Neural Network: Application to Program
  Reidentification},'' in \emph{Proceedings of the 2019 SIAM International
  Conference on Data Mining}.\hskip 1em plus 0.5em minus 0.4em\relax SIAM,
  2019, pp. 693--701.

\bibitem{mucha2010community}
P.~J. Mucha, T.~Richardson, K.~Macon, M.~A. Porter, and J.-P. Onnela,
  ``{Community Structure in Time-Dependent, Multiscale, and Multiplex
  Networks},'' \emph{science}, vol. 328, no. 5980, pp. 876--878, 2010.

\bibitem{gehring2017convolutional}
J.~Gehring, M.~Auli, D.~Grangier, D.~Yarats, and Y.~N. Dauphin,
  ``{Convolutional Sequence to Sequence Learning},'' in \emph{International
  conference on machine learning}.\hskip 1em plus 0.5em minus 0.4em\relax PMLR,
  2017, pp. 1243--1252.

\bibitem{battaglia2018relational}
P.~W. Battaglia, J.~B. Hamrick, V.~Bapst, A.~Sanchez-Gonzalez, V.~Zambaldi,
  M.~Malinowski, A.~Tacchetti, D.~Raposo, A.~Santoro, R.~Faulkner
  \emph{et~al.}, ``Relational inductive biases, deep learning, and graph
  networks,'' \emph{arXiv preprint arXiv:1806.01261}, 2018.

\bibitem{de2005tutorial}
P.-T. De~Boer, D.~P. Kroese, S.~Mannor, and R.~Y. Rubinstein, ``{A Tutorial on
  the Cross-Entropy Method},'' \emph{Annals of Operations Research}, vol. 134,
  pp. 19--67, 2005.

\bibitem{RobotCarDatasetIJRR}
W.~Maddern, G.~Pascoe, C.~Linegar, and P.~Newman, ``{1 Year, 1000km: The Oxford
  RobotCar Dataset},'' \emph{The International Journal of Robotics Research
  (IJRR)}, vol.~36.

\bibitem{Kim2020MulRanMR}
G.~Kim, Y.-S. Park, Y.~Cho, J.~Jeong, and A.~Kim, ``{MulRan: Multimodal Range
  Dataset for Urban Place Recognition},'' \emph{2020 IEEE International
  Conference on Robotics and Automation (ICRA)}, pp. 6246--6253, 2020.

\bibitem{hui2022efficient}
L.~Hui, M.~Cheng, J.~Xie, J.~Yang, and M.-M. Cheng, ``{Efficient 3D Point Cloud
  Feature Learning for Large-Scale Place Recognition },'' \emph{IEEE
  Transactions on Image Processing}, vol.~31, pp. 1258--1270, 2022.

\bibitem{hou2022hitpr}
Z.~Hou, Y.~Yan, C.~Xu, and H.~Kong, ``{HiTPR: Hierarchical Transformer for
  Place Recognition in Point Cloud},'' in \emph{2022 International Conference
  on Robotics and Automation (ICRA)}.\hskip 1em plus 0.5em minus 0.4em\relax
  IEEE, 2022, pp. 2612--2618.

\bibitem{zhou2021ndt}
Z.~Zhou, C.~Zhao, D.~Adolfsson, S.~Su, Y.~Gao, T.~Duckett, and L.~Sun,
  ``{\textit{NDT-Transformer}: Large-Scale 3D Point Cloud Localisation using
  the Normal Distribution Transform Representation},'' in \emph{2021 IEEE
  International Conference on Robotics and Automation (ICRA)}.\hskip 1em plus
  0.5em minus 0.4em\relax IEEE, 2021, pp. 5654--5660.

\bibitem{fan2022svt}
Z.~Fan, Z.~Song, H.~Liu, Z.~Lu, J.~He, and X.~Du, ``{SVT-Net: Super
  Light-Weight Sparse Voxel Transformer for Large Scale Place Recognition},''
  in \emph{Proceedings of the AAAI Conference on Artificial Intelligence},
  vol.~36, no.~1, 2022, pp. 551--560.

\bibitem{xia2022pvt3d}
Y.~Xia, M.~Gladkova, R.~Wang, J.~F. Henriques, D.~Cremers, and U.~Stilla,
  ``{PVT3D: Point Voxel Transformers for Place Recognition from Sparse Lidar
  Scans},'' \emph{arXiv preprint arXiv:2211.12542}, 2022.

\end{thebibliography}

\end{document}